\pdfoutput=1

\documentclass[11pt]{article}

\usepackage[]{emnlp2021}

\usepackage{times}
\usepackage{latexsym}

\usepackage[T1]{fontenc}

\usepackage[utf8]{inputenc}

\usepackage{microtype}

\usepackage{booktabs}
\usepackage{kotex}
\usepackage{graphicx}
\usepackage{longtable}
\usepackage{bbold}
\usepackage{amsmath}
\usepackage{cleveref}
\usepackage{multirow}
\usepackage{multicol}
\usepackage{soul}

\usepackage{array}

\newcommand{\congress}{the 111\textsuperscript{th} U.S. Congress}

\title{Learning Bill Similarity with Annotated and Augmented Corpora of Bills}

\author{Jiseon Kim$^{\dag}$, Elden Griggs$^{\S}$, In Song Kim$^{\S}$, Alice Oh$^{\dag}$ \\
  $^{\dag}$School of Computing, KAIST  \\
  $^{\S}$Department of Political Science, MIT \\
  \texttt{jiseon\_kim@kaist.ac.kr} \\
  \texttt{\{emgriggs, insong\}@mit.edu} \\
  \texttt{alice.oh@kaist.edu} \\}

\begin{document}
\maketitle
\begin{abstract}
Bill writing is a critical element of representative democracy. However, it is often overlooked that most legislative bills are derived, or even directly copied, from other bills. Despite the significance of bill-to-bill linkages for understanding the legislative process, existing approaches fail to address semantic similarities across bills, let alone reordering or paraphrasing which are prevalent in legal document writing. In this paper, we overcome these limitations by proposing a 5-class classification task that closely reflects the nature of the bill generation process. In doing so, we construct a human-labeled dataset of 4,721 bill-to-bill relationships at the subsection-level and release this annotated dataset to the research community. To augment the dataset, we generate synthetic data with varying degrees of similarity, mimicking the complex bill writing process. We use BERT variants and apply multi-stage training, sequentially fine-tuning our models with synthetic and human-labeled datasets. We find that the predictive performance significantly improves when training with both human-labeled and synthetic data. Finally, we apply our trained model to infer section- and bill-level similarities. Our analysis shows that the proposed methodology successfully captures the similarities across legal documents at various levels of aggregation. \footnote{We release our human-annotated \& synthetic dataset and code at \url{https://github.com/hikoseon12/learning-bill-similarity}}

\end{abstract}

\section{Introduction}
\begin{figure*}[!t]
\centering
\includegraphics[width=1.\textwidth]{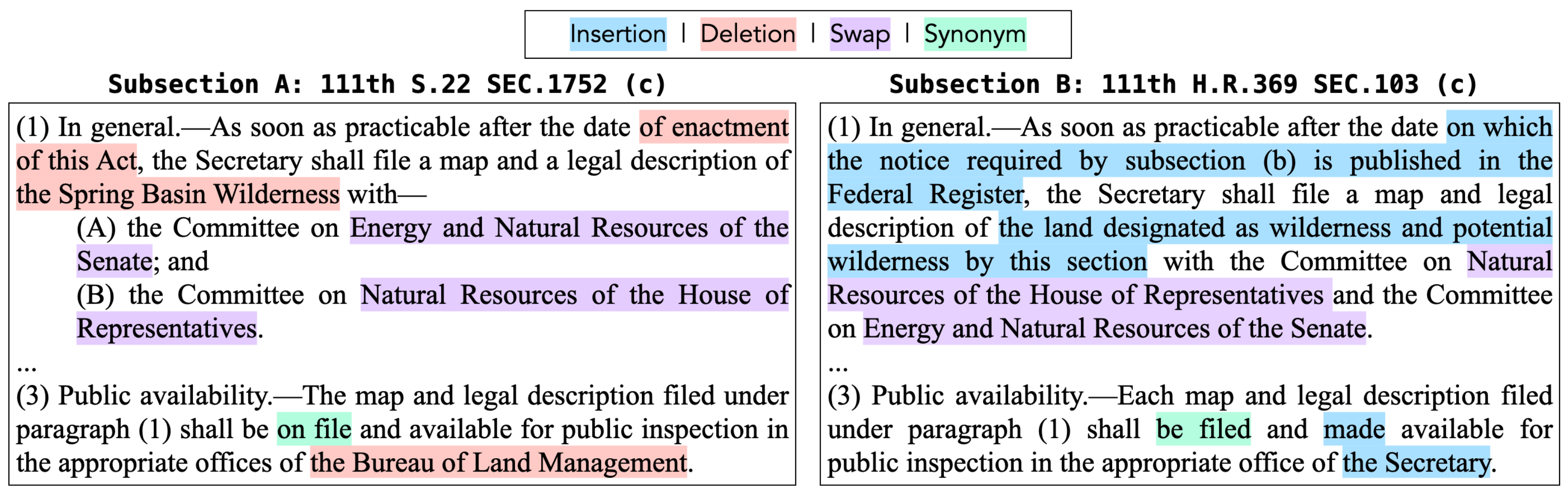}
\caption[]{An example of a subsection pair sharing similar policy ideas. We highlight the different parts. This example shows the possible text edit operations: insertion, deletion, swap, and synonym replacement.}
\label{intro:subsection_example}
\end{figure*}

In the U.S. and other democracies, a bill becomes a law through a complex political process, which includes a series of political debates and compromises within their legislative branch. A consistent pattern that arises from this political interaction is the emergence of several lengthy omnibus bills which cover a range of topics/special interest and combine parts of numerous smaller bills. In fact, most bills tend to be a mixture of a number of other bills, whether enacted or ``dead.'' Furthermore, in the U.S., many bills are modified and re-introduced between the two chambers of Congress. Uncovering these complex relationships among bills is important because it helps scholars better understand the legislative process, one of the most important parts of representative democracy. In this paper, we develop a methodology to characterize this relationship across bills. Specifically, we leverage semantic similarities between subsections to infer bill linkages using annotated and augmented synthetic corpora.

Previous studies have investigated this issue in both political science~\citep{ wilkerson2015tracing,linder2020text} and computer science~\citep{burgess2016prototype, burgess2016legislative}. To find similar bills, they use either a simple bag-of-words model or the Smith-Waterman local alignment algorithm (SWAlign)~\citep{gotoh1982improved} . Although this latter approach is useful in identifying exact text matches, it cannot capture semantic similarities across bills that arise from sentence reordering or paraphrasing. Figure~\ref{intro:subsection_example} shows two closely related bills that contain nearly identical policy ideas, but this similarity would be difficult to detect with SWAlign. We note the differences between these bills arise primarily from their distinct bill structures and insertion and deletion of words and sentences.

We propose a 5-class classification task that closely reflects the nature of the bill generation process. We begin by constructing a novel human-labeled dataset of  4,721 bill-to-bill relationships across {\congress} bills at the subsection level. Using this data, we apply multi-stage training and perform two-stage fine-tuning on models for classification. Specifically, we train several BERT variants including 
BERT~\citep{devlin-etal-2019-bert}, RoBERTa~\citep{liu2019roberta}, and LEGAL-BERT~\citep{chalkidis-etal-2020-legal}.
Our analysis shows that the two-stage training significantly improves the classification performance across all models. Finally, we extend the trained model to section- and bill-level pairs, which confirms that our methodology successfully determines the similarity at higher levels of aggregation between legal documents. In sum, our contributions are as follows: 
\begin{itemize}
\item We define a task of subsection-level legislative text similarity and publish a human-annotated dataset of 4,721 subsection pairs.
\item We describe and share a corpus of synthetic subsection pairs that reflects the bill modification process and improves the subsection-level classification.
\item We extend our subsection-level approach to inferring the pairwise similarity at both the section and bill level.
\end{itemize}

\section{Related Work}
\paragraph{Bill Similarity}
Several studies have examined bill linkages.
\citet{burgess2016prototype} measure sentence pair similarity in state legislation using a  unigram bag-of-words model. 
Others use the SWAlign algorithm to detect aligned sequences of text between bills~\citep{burgess2016legislative, linder2020text, wilkerson2015tracing}.
In contrast, we use Transformer-based language models (RoBERTa, Legal-BERT, and BERT) trained with human-annotated and synthetically generated bill pairs.

\paragraph{Citation Recommendation}
Citation recommendation involves finding relationships between document-sized texts such as papers or articles \citep{bhagavatula-etal-2018-content, jyunyu2019semantic}.
A bill  is also in the form of a document and bill similarity can be posed as a task to retrieve related legislation. Recently, this is done with ensemble models of traditional information retrieval (IR) models and BERT variants for pre-fetching and re-ranking in \citet{chalkidis-etal-2021-regulatory}. 
One weakness of their study is that they rely only on the obvious relationship between an EU directive and the corresponding UK regulation that complies with the directive, ignoring many other legal text pairs that may be related in other ways. 
To overcome this weakness, we find the linkage of bills in a bottom-up manner by splitting a bill into subsection pieces. Subsections are smaller units and are easier to both annotate and measure similarity. From subsection results, we aggregate similarity scores up to the bill-level.
We engage in large-scale human-annotation of legal text similarities. We are, in fact, working with magnitudes larger set of bills, as most bills are introduced but do not become law, and many are re-introduced with modifications.

\paragraph{Plagiarism Detection}  
Plagiarism detection is also a task of finding a connection between two texts. Work done by \citet{potthast2011overview} in plagiarism detection involves detecting similar plagiarized parts between two documents. While this method could be useful to determining similarities between bills, it is difficult to indicate the degree of similarity between two texts using this method. Additionally, the dataset is collected by crowdsourcing with multiple topics, which means the data does not reflect the characteristics of bills. In this paper, we propose a new dataset that represents the similarity of bill pairs by reflecting the unique characteristics of legislation.

\paragraph{Multi-stage Training}
Multi-stage training improves downstream tasks, especially for a small dataset or a domain-specific task~\citep{gururangan-etal-2020-dont, pruksachatkun-etal-2020-jiant}. There are many different BERT models pretrained for various domains and tasks~\citep{chalkidis-etal-2020-legal, beltagy-etal-2019-scibert,  10.1093/bioinformatics/btz682, feng-etal-2020-codebert}. These models are further trained with unlabeled large domain-specific corpora or synthetic training data~\citep{zhang-etal-2020-multi-stage}.
The subsection pair dataset we build is relatively small. To improve the classification performance, therefore, we apply two-stage training method with bills as text corpus. 
In doing so, we build a large synthetic pairs dataset and perform two-stage fine-tuning on models.

\paragraph{Data Augmentation}  
There are several data augmentation techniques used in NLP~\citep{sennrich-etal-2016-improving, chen-etal-2020-mixtext, wei-zou-2019-eda}. Among them, we apply EDA~\citep{wei-zou-2019-eda} to produce synthetic subsection pairs because EDA can effectively control the degree of similarity with word-based modifications. We apply the four operations of EDA, but we slightly modify the replacement operation to replace with random sequences from one of the bills in the  dataset~(\ref{generation:synthetic_pairs}).

\section{Annotating Subsections}
\subsection{Subsection Units}
A bill consists of a series of smaller units---e.g., chapters, sections, and subsections---concerning diverse policy ideas. To facilitate the comparison across lengthy bills, we take the subsection as the basic unit of analysis. 
Subsections are self-sufficient units in bills in that they are at least a sentence and are grammatically complete ~\citep{strokoff2007legislative}.  
Moreover, our choice of subsection as a unit of analysis can be further justified by the drafting style guidelines present in the U.S. whereby subsections are described as the
smallest common unit of legislation across styles ~\citep{strokoff2007legislative}. Note that we later aggregate the subsection level similarities to determine  section and bill pair similarity in Section~\ref{application}.

\subsection{Subsection Relation Types}\label{sec:measure}
\label{bill_relation:subsection_relation_type}
We propose the following five-class classification to capture subsection similarities distinguishing word-level and policy idea changes.

\smallskip
\noindent \textbf{[4] Identical} Two subsections are identical.

\noindent \textbf{[3] Almost Identical} Two subsections share identical text, but contain minor changes such as paraphrasing and single word changes, so that the legislative idea is almost equivalent. We distinguish this label from ``Identical'' because even a single word change could lead to a different policy idea (e.g., \textit{tax credit} vs. \textit{tax exemption}).

\noindent \textbf{[2] Related} Two subsections share more than 50\% of legislative text, not considering common legislative phrases. This class implies that, even at the granular subsection level, one or two distinct policy ideas may be added or deleted from another bill.

\noindent \textbf{[1] Partially Related} Two subsections share less than 50\%  of legislative text, not considering common legislative phrases. This class implies that two subsections might only share 1 or 2 policy ideas while differing in all other ways. 

\noindent \textbf{[0] Unrelated} Two subsections do not share any policy ideas.

We provide detailed descriptions for each class in Table \ref{appendix:desciprtion_coding_rules} in Appendix \ref{appendix:coding_rules}.

\section{Data Generation}
\begin{table}[!tb]
\centering
\small
\renewcommand{\arraystretch}{1.2}
\begin{tabular}{@{}l|r@{}}
\toprule
Text            & \#             \\ \midrule
Bills           & 10,621             \\
Subsections     & 175,736            \\
Subsections without boilerplates & \textbf{126,541} \\ 
\midrule
Average \# words in subsection & 154.6            \\
\bottomrule
\end{tabular}
\caption{Text data from \congress.}
 \label{data_generation:bill_stats}
\end{table}

\begin{table}[!t]
\centering
\small
\begin{tabular}{@{}l|r|rrrr@{}}
\toprule
\multirow{2}{*}{Class} & \multicolumn{1}{c|}{\multirow{2}{*}{Size}} & \multicolumn{1}{c}{\multirow{2}{*}{Cons.}} & \multicolumn{3}{c}{\# Coders} \\ \cmidrule{4-6}
                       & \multicolumn{1}{c|}{}                      & \multicolumn{1}{c}{}                      & 2          & 3       & 4      \\ \midrule
4 Identical            & 801                                       & 39                                        & 757        & 4       & 1      \\
3 Almost Identical     & 524                                       & 21                                        & 446        & 54      & 3      \\
2 Related              & 679                                       & 28                                        & 439        & 205     & 7      \\
1 Partially Related    & 300                                       & 8                                         & 155        & 128     & 9      \\
0 Unrelated            & 2,417                                     & 54                                        & 2,259      & 103     & 1      \\ \midrule
Total                  & 4,721                                     & 150                                       & 4,056      & 494     & 21     \\ \bottomrule
\end{tabular}
\caption{The agreement statistics of the dataset. \textit{Size} is the total number of human annotated pairs for each class. The \textit{Cons.} column indicates the number of data used to help annotators reach the consensus. \textit{\#~Coders} is the number of annotators needed to reach the final consensus label.}
\label{data_generation:diff_answer}
\end{table}

\begin{figure*}[!t]
\centering
\includegraphics[width=1.\textwidth]{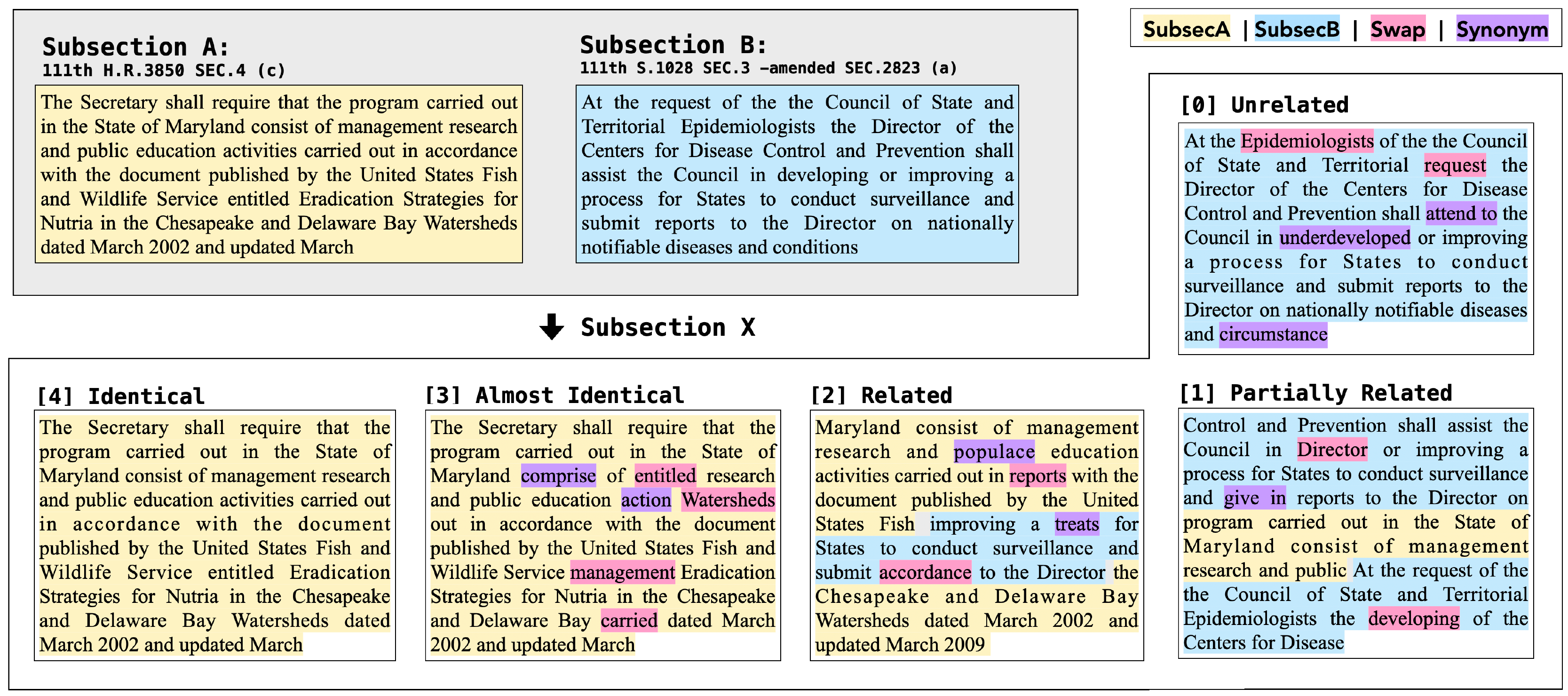}
\caption[An example of a subsection pair]{Synthetic subsection pair generation. We generate Subsection A's relation type pair, Subsection X by (1) deleting phrases in Subsection A, (2) inserting phrases from randomly selected Subsection B, and (3) applying swap and synonym operations. The final output of synthetic subsection pair is (Section A, Section X).}
\label{generation:subsectionx}
\end{figure*}

This section explains how we annotate subsection pairs and generate the synthetic dataset.

\subsection{Preprocessing}
We investigate the linkages across the House and the Senate bills (prefixed with H.R. and S., respectively) from~\congress~(2009-2010).\footnote{Data from \url{https://github.com/unitedstates/congress}} As shown in Table~\ref{data_generation:bill_stats}, ~\congress~has 10,621 bills. We focus our analysis on the latest version of each bill. After removing all non-alphanumeric symbols and tables from the texts, we split a bill into subsection level pieces. When a subsection quotes another bill, the quoted portion is split into a separate piece to properly capture distinct policy ideas. This yields 175,736 subsections.

To exclude commonly used boilerplate, we manually build a list of common section headings associated with boilerplate~(Table \ref{appendix:boilerplate_example} in Appendix \ref{appendix:boilerplate}) and remove subsections with these headings.
This removes 26,055 subsections from the dataset.
A subsection of 30 words or less is excluded as it is difficult to measure the similarity for those cases, while a subsection with a length longer than 512 tokens is sliced into smaller text pieces each having 400 words at maximum. 
As a result, the dataset after preprocessing consists of 126,541 subsections that are 154.6 words long on average (see Table~\ref{data_generation:bill_stats}).

\subsection{Human Annotated Pairs}
With preprocessed subsection text, we build the ground truth of subsection pairs. Since most subsection pairs are likely to be \textit{Unrelated}, we sample subsection pairs that exhibit high similarity between their embeddings to balance the annotated dataset. Specifically, we compute the cosine similarity score for all subsection pairs using Sentence-BERT embeddings pre-trained on NLI datasets~\citep{reimers-gurevych-2019-sentence}. Among the subsection pairs with cosine similarity scores between 0.85 and 1.0\footnote{Our model also performs well outside of the set of pairs (pairs with similarity less than 0.85). Further experiments with different pairs can be found in Appendix \ref{appendix:extra_annotation_experiment}.}, we randomly sample 5,400 pairs and ask annotators to label them.

We recruit nine annotators, including two of the co-authors. Before allocating the annotation task to individuals, we first ask them to reach agreement in labeling based on Coding Rules shown in Table \ref{appendix:desciprtion_coding_rules} in Appendix \ref{appendix:coding_rules}. For 200 instances, they come to a total consensus.
After the consensus stage, we have pairs of coders annotate sets of subsection pairs, using the Coding Rules.  
After this step, if the labels between annotators differ, we recruit another annotator to decide between the two labels (\textit{\# Coders 3} in Table ~\ref{data_generation:diff_answer}). If the annotators still cannot come to a conclusion, the authors make the final decision (21 pairs, \textit{\# Coders 4} in Table~\ref{data_generation:diff_answer}).

We remove pairs if at least one annotator marked it as boilerplate. After excluding boilerplate, we have 4,721 human-annotated pairs in the dataset. The average Cohen's Kappa score~\citep{cohen1968weighted}~is 0.807 (0.764 if boilerplate pairs are included).
As shown in  Table~\ref{data_generation:diff_answer}, the \textit{Unrelated} class accounts for the largest portion of annotated pairs at 51.2\% followed by \textit{Identical, Related, Almost Identical}, and \textit{Partially Related}. The \textit{Partially Related} class has the lowest agreement between annotators.

\subsection{Synthetic Pairs}
\label{generation:synthetic_pairs}

Figure~\ref{generation:subsectionx} illustrates how we generate a synthetic pair: \textit{Subsection A} and its counterpart, \textit{Subsection X} for each relation type. We apply EDA~\citep{wei-zou-2019-eda} modifying it to fit the subsection relation types. EDA has four techniques for generating a training set: random swap, synonym replacement, random deletion, and random insertion.
For~\textit{\textbf{swap}}, the two randomly selected words are switched with each other. For~\textit{\textbf{synonym}}, a randomly picked word, stopwords excluded, is replaced with its synonym, using WordNet~\citep{miller1995wordnet}.
For~\textit{\textbf{deletion}} and ~\textit{\textbf{insertion}}, we modify word-level to n-gram-level operation reflecting the nature of bill generation.
With \textit{\textbf{deletion}}, we select n-grams chunks to be deleted from \textit{Subsection A}.
Finally, we select n-grams chunks to be added to \textit{Subsection X} from a randomly selected \textit{Subsection B}. The following provides further details of the generation of each synthetic pair type:
\smallskip

\noindent \textbf{[4] Identical}
\textit{Subsection X} has exactly the same content with \textit{Subsection A}.

\noindent \textbf{[3] Almost Identical}
We apply ~\textit{\textbf{swap}} and~\textit{\textbf{synonym}} operations to \textit{Subsection A} for between 0 to 10\% of \textit{Subsection A}'s  length. We empirically set the maximum number of applying the two operations as 20.

\noindent \textbf{[2] Related}
In addition to~\textit{\textbf{swap}}~and~\textit{\textbf{synonym}} operations, we apply \textit{\textbf{insertion}} and \textit{\textbf{deletion}}  as shown in Figure~\ref{generation:subsectionx}. 
With ~\textit{\textbf{insertion}} and \textit{\textbf{deletion}}, we randomly select n-gram chunks, setting the proportion of modification to between 20\% and 40\% of \textit{Subsection A}'s length. The proportion of modification and the ratio of \textit{\textbf{insertion}} and \textit{\textbf{deletion}} are all randomly chosen.
We put the n-gram chunks together in random order to generate \textit{Subsection X}. \textit{\textbf{swap}}~and~\textit{\textbf{synonym}} are applied similar to how a \textit{Almost Identical} pair is generated. 

\noindent \textbf{[1] Partially Related}
These are generated in a similar manner to the \textit{Related} class, but the only difference is that the proportion of modifications is set larger to be between 60\% to 80\%.

\noindent \textbf{[0] Unrelated}
We randomly select \textit{Subsection X} from the subsection pool and also apply~\textit{\textbf{swap}}~and~\textit{\textbf{synonym}} operations.

\section{Experiment Setting}
We conduct subsection-level classification tasks with various settings, four different models and multiple sizes of combination for both synthetic and human-annotated datasets. 

\subsection{Dataset}
We use both human-labeled and synthetic datasets for training.
The human-annotated data instances are split into train, validation, and test set with the ratio of 7:1:2 (3305, 472, and 944, respectively). 
To prevent imbalanced-class pairs when using the relatively small size of human-labeled pairs for training, we try to balance the classes in the dataset as much as possible.
We use synthetic data for training only. The size of synthetic pairs differs depending on experiment settings. The number of instances over each class is identical.

\subsection{Models}
We use three BERT models and the SWAlign alignment algorithm~\citep{gotoh1982improved}. 

\noindent \textbf{BERT models} For BERT~\citep{devlin-etal-2019-bert}, LEGAL-BERT~\citep{chalkidis-etal-2020-legal}, and RoBERTa~\citep{liu2019roberta}, we feed [CLS] representation into the final layer for 5-class subsection-level classification following standard practice~\citep{devlin-etal-2019-bert}.

\noindent \textbf{SWAlign} We use SWAlign with logistic regression. The SWAlign algorithm calculates the similarity score given two input subsections. With the similarity scores, logistic regression classifies a subsection pair among 5 classes.
We set the detail settings of SWAlign method to be the same as those given in the previous study~\citep{wilkerson2015tracing}. 

\subsection{Experiment Details}
We implement models using the PyTorch version of \textit{BERT-base-cased\footnote{\url{https://huggingface.co/bert-base-cased}}, LEGAL-BERT-base-uncased\footnote{\url{https://huggingface.co/nlpaueb/legal-bert-base-uncased}}, RoBERTa-base\footnote{\url{https://huggingface.co/roberta-base}}} from Huggingface.
The hyperparameter tuning uses the following ranges: a learning rate $\in \{1e-05, 2e-05, 3e-05\}$, number of training epochs $\in \{2, 3, 4\}$. Batch size is 32 and the maximum epoch is 4.
We set optimal learning rates for each model with 10,000 updates of warm-up for AdamW optimizer.
The score of all experiment results is the average of five trials, except ablation study (\ref{result:ablation}) score, which is the average of 10 trials because of the large variance of scores.
The best model parameter settings are in Appendix \ref{appendix:model_settings}.

\section{Results}
\begin{figure}[!t]
\centering
\small
\includegraphics[width=0.45\textwidth]{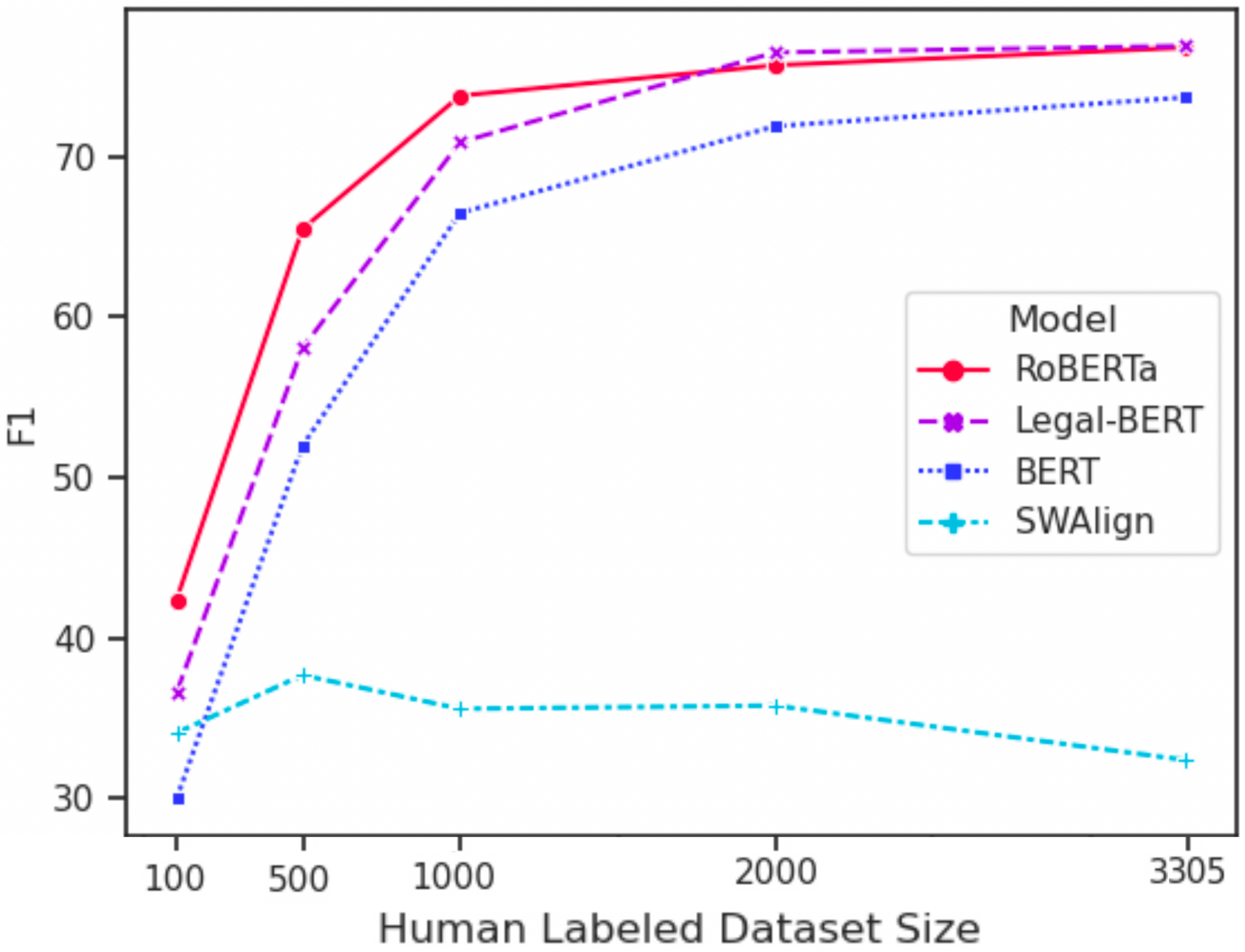}

\caption[]{Macro F1 score depending on model and human-labeled data size.}
\label{result:human_model_f1}
\end{figure}

We compare the performance of the models with various experiment settings described above.

\subsection{Classification Results Trained on Human-labeled Dataset}
Figure~\ref{result:human_model_f1} shows the performance of four models trained on the different sizes of the human-labeled dataset. For language models (RoBERTa, Legal-BERT, and BERT), their F1 scores increase as the size of the dataset grows, whereas the score of SWAlign is invariant to the size.
Overall, the language models show better performance than SWAlign. Trained on the whole dataset, RoBERTa (76.8) and Legal-BERT (76.9) show the highest performance, followed by BERT (73.7) and SWAlign (37.7, trained on 500 pairs). The superior performance of Legal-BERT compared to BERT shows that pre-training on domain-specific knowledge helps to predict the subsection-level similarity.

\subsection{Ablation Study of Synthetic Dataset}
\label{result:ablation}
To demonstrate the performance of augmentation operations, we conduct an ablation study for the synthetic pairs. We compare the results of three synthetic pairs: swap only, synonym only, and then both swap and synonym. The size of all the training synthetic pairs is set as 10,000 pairs—the results across the language models are shown in Table~\ref{result:ablation_test}. 
As the table suggests, applying both swap and synonym operations helps to increase the classification performance overall.

\begin{table}[!t]
\centering
\tiny
\begin{tabular}{@{}l|cc|cc|cc@{}}
\toprule
          & \multicolumn{2}{c|}{RoBERTa}   & \multicolumn{2}{c|}{Legal-BERT} & \multicolumn{2}{c}{BERT}      \\
Operation & Acc.          & F1            & Acc.           & F1            & Acc.          & F1            \\ \midrule
swap      & 70.7$_{2.9}$          & 62.6$_{2.2}$          & 62.1$_{2.0}$           & 55.6$_{1.6}$          & 58.5$_{3.4}$          & 51.2$_{3.3}$          \\
synonym   & 68.6$_{2.5}$          & 61.7$_{1.7}$          & 63.0$_{1.5}$           & \textbf{57.2}$_{1.4}$ & 58.4$_{3.1}$          & \textbf{52.4}$_{2.2}$ \\
both      & \textbf{73.5}$_{2.6}$ & \textbf{64.0}$_{2.1}$ & \textbf{63.8}$_{2.9}$  & 56.4$_{1.6}$          & \textbf{58.7}$_{2.2}$ & 52.3$_{1.5}$          \\ \bottomrule
\end{tabular}
\caption{Ablation study of augmentation operations: swap and synonym. Results are from models trained on 10,000 synthetic pairs.}
\label{result:ablation_test}
\end{table}

\begin{table}[!t]
\centering
\small
\begin{tabular}{@{}l|ccc@{}}
\toprule
Class               & Synthetic & Human & Syn.+Human \\ \midrule
4 Identical         & 92.2$_{3.3}$      & 95.6$_{1.2}$        & \textbf{96.9}$_{0.3}$   \\
3 Almost Identical  & 63.4$_{4.5}$      & 74.7$_{8.1}$        & \textbf{77.6}$_{4.8}$   \\
2 Related           & 62.6$_{5.6}$      & 72.6$_{2.6}$        & \textbf{76.3}$_{1.2}$   \\
1 Partially Related & 17.7$_{4.0}$      & 45.5$_{3.2}$        & \textbf{51.9$_{3.1}$}   \\
0 Unrelated         & 84.2$_{2.2}$      & 95.8$_{0.8}$        & \textbf{97.1}$_{0.7}$   \\ \midrule
Average Accuracy    & 73.5$_{2.6}$      & 86.9$_{1.0}$        & \textbf{88.9}$_{0.8}$        \\
Average Macro F1    & 64.0$_{2.1}$       & 76.8$_{1.8}$       & \textbf{79.9}$_{1.4}$        \\ \bottomrule
\end{tabular}
\caption[]{F1 of each class, average accuracy and average F1 for three different training data types: synthetic, human-labeled, synthetic+human-labeled dataset. We report results generated from RoBERTa.}
\label{result:f1_score_class_test}
\end{table}

\subsection{Effectiveness of Synthetic Dataset}
Table~\ref{result:f1_score_class_test} shows the effectiveness of the synthetic dataset, comparing the F1 score of 5-class subsection similarity classification on different training data types. The \textit{Synthetic} and \textit{Human} columns are the results of models trained on synthetic and human-labeled datasets, respectively. The \textit{Synthetic+Human} column is the result of a two-stage training setting. We train models on the synthetic dataset first, and then re-train models on the human-labeled set. We report RoBERTa results, and use 10,000 synthetic pairs and 3,305 human-labeled pairs.
We find that training on synthetic pairs only results in an accuracy score of 73.5 and an F1 score of 64.0. 
Multi-stage training with both synthetic and human-annotated data significantly improves performance for all classes, increasing the average accuracy by 2.0 points and the F1 score by 3.1 points versus training only with human-annotated data.
Note that the \textit{Partially Related} class, which is the smallest and has the lowest consensus among the 5 classes, has the lowest score for all three settings. In this case, the two-stage training leads to the largest increase (+6.4) in performance. This result suggests that synthetic pairs most significantly supplement the performance of the smallest and most difficult class.

\begin{figure}[!t]
\centering
\small
\includegraphics[width=0.45\textwidth]{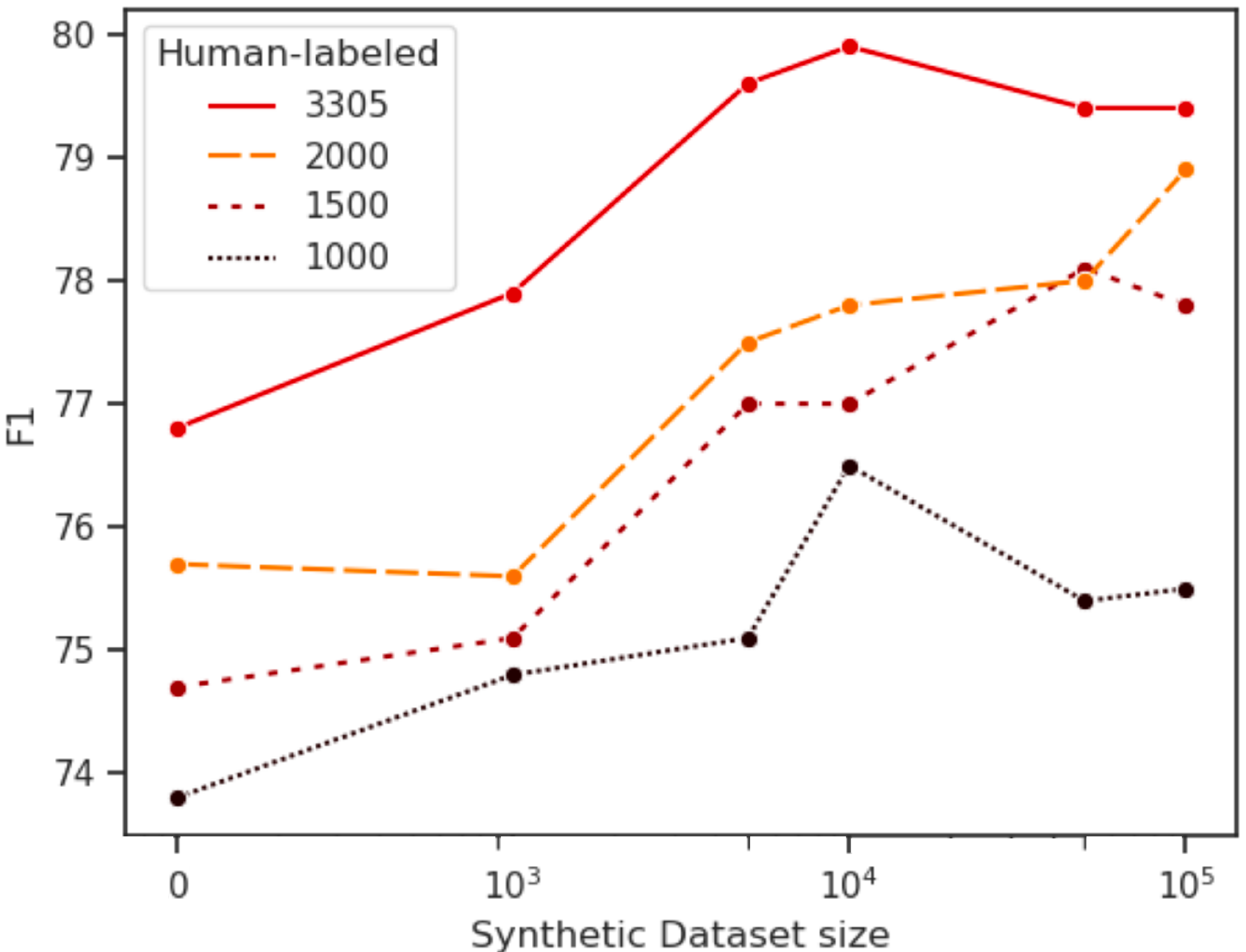}
\caption[]{Macro F1 score of RoBERTa depending on the size of human-labeled and synthetic pairs.}
\label{result:roberta_f1}
\end{figure}

\subsection{Performance for Different Size of Human-labeled and Synthetic Dataset}
We vary the size of the human-annotated dataset, as well as the size of the synthetic dataset. Note that collecting human labels is costly, especially in this domain where the annotators need to be sufficiently trained about legislation text. Generating synthetic data is relatively simple, so the size of the synthetic dataset is limited only by the computational resources.
We apply two-stage training with various sizes of synthetic and human-labeled data. We run multiple experiments for all possible combinations of \textit{(\#synthetic, \#human)}, where \textit{\#synthetic $\in$} \{0, 1K, 5K, 10K, 20K, 50K, 100K\} and \textit{\#human $\in$} \{1000, 1500, 2000, 3305\}.

Overall, the score of all models tend to increase when the size of synthetic set grows. RoBERTa achieves the highest performance among the models, so we show the RoBERTa results in  Figure~\ref{result:roberta_f1}. Results with other models are in Table \ref{appendix:classification_test} Appendix \ref{appendix:result}. Generally, the scores tend to increase when increasing both human-labeled and synthetic dataset size.
When using 3,305 human-labeled pairs, the F1 score plateaus around 10K synthetic pairs.

To see what would happen with a limited number of human-labeled pairs, we train with 1000, 1500, and 2000 human-annotated pairs. The results show that having more human-annotated data is always helpful, but with a smaller dataset of 2000 pairs the final performance with 100,000 synthetic pairs comes quite close to the results using the entire human-annotated dataset. Further, even with less than half at just 1500, using the 100,000 synthetic pairs improves performance over using just 3,305 human-annotated pairs. This result suggests the synthetic dataset effectively boosts the classification performance and can be used as a partial substitute for the costly human-labeled data.


\section{Application}
\label{application}
We evaluate the performance of our trained model at two distinct levels of aggregation: (1) section-level and (2) bill-level. Our analysis is motivated by the  legislative process through which political interests at the subsection-level could be aggregated into larger legislation as politicians engage in negotiations and logrolling. Specifically, we develop a measure of text similarity at higher levels of aggregation. We evaluate the performance of our measurements against the section-level similarity scores developed by \citet{wilkerson2015tracing} and the proxy bill-level similarity measure based on the co-occurrence of lobbying activities. We compare the results with RoBERTa and SWAlign that were trained with the best performance setting: 10,000 synthetic pairs and 3,305 human-labeled pairs for RoBERTa, 1,000 synthetic pairs and all 3,305 the human-labeled pairs for SWAlign.

\subsection{Section-level Similarity}
\citet{wilkerson2015tracing} developed 6 alignment categories at the section-pair level according to the similarity in policy ideas: \textit{1: Clear Policy Idea Match}, \textit{2: Likely Policy Idea Match}, \textit{3: Same Topic, Different Idea}, \textit{4: Different Topic}, \textit{5. Boilerplate}, and \textit{6: Junk}. They focus on the similarity between the sections from bills that are related to the Patient Protection and Affortable Care Act (PPACA), known as Obamacare (H.R.3590) in~\congress.

To facilitate a direct comparison between their hand-coded similarity measures at the section-level and ours, we mapped our 5 subsection relation types to their 4 section-level alignment types while discarding the ``Boilerplate'' and ``Junk'' categories.  This yields the following four categories: (a) <\textit{Identical} and \textit{Almost Identical}> to <\textit{Clear Policy Idea Match}>, (b) <\textit{Related}> to <\textit{Likely Policy Idea Match}>, (c) <\textit{Partially Related}> to <\textit{Same Topic, Different Idea}>, and (d) <\textit{Unrelated}> to <\textit{Different Topic}>. We focus our analysis on 1,091 section pairs after excluding 127 outlier section pairs with longer text length.

\begin{table}[!t]
\small
\centering
\begin{tabular}{@{}l|cc@{}}
\toprule
Model               & SWAlign & RoBERTa       \\ \midrule
Accuracy            & 42.8    & \textbf{52.4} \\
macro F1            & 39.8    & \textbf{45.3} \\ 
Pearson correlation & 57.5    & \textbf{58.2} \\
\bottomrule
\end{tabular}
\caption[]{Section-level correlation/classification results.}
\label{application:section_result}
\end{table}

\paragraph{Section-level Results.} Table~\ref{application:section_result} presents the section-level result.
Across all metrics (Accuracy, macro F1, and Pearson correlation), RoBERTa outperforms SWAlign. Specifically, there is a significant gap in the classification result (Acc. and F1) whereby RoBERTa gets a higher score and better represents the degree of similarity compared to SWAlign. Notice that, despite the significant difference in their predictive performance (p-value from t-test: 0.0038 < 0.01, **), both models return highly correlated results (SWAlign: 57.5, RoBERTa: 58.2). This is because SWAlign returns many inaccurate predictions across the classes, but the predicted categories still follow the general numeric ordering ranging from Different Topic class to Clear Policy Idea Match class. In contrast, RoBERTa returns more accurate predictions across the categories. Our findings show that the models with the subsections and five categories perform well even when the predictive task is slightly adjusted. We also suggest that higher performance at the granular subsection level can be extended to the better predictive performance at the section-level.

\subsection{Bill-level Similarity}
We extend our analysis to the bill-level in order to investigate whether the performance of the trained model could further be extended to higher-level aggregation. Our analysis fills an important gap in empirical research in political science where scholars are limited in their understanding of the relations across congressional bills.

\paragraph{Co-occurence of Bills in Lobbying Reports.} To evaluate the performance of the trained model, we begin by developing a proxy measure of bill-level similarity based on the co-occurrence of any pair of bills in lobbying reports using data from LobbyView \footnote{\url{https://www.lobbyview.org}} \citep[see][for details]{kim:kunisky:20}. We then count the number of times any two given pair of bills appear together within the same lobbying report and within the same issue category (e.g., technology). To be sure, two unrelated bills could be reported together under the same issue section. Nonetheless, two bills are likely to be related if they are reported to be lobbied together multiple times across numerous reports filed by distinct interest groups. For instance, Facebook reported to have lobbied on two related bills---CARES Act (PL 116-136) and Families First Coronavirus Response Act (PL 116-127)---along with many other interest groups.

\begin{figure}[!t]
\centering
\includegraphics[width=0.45\textwidth]{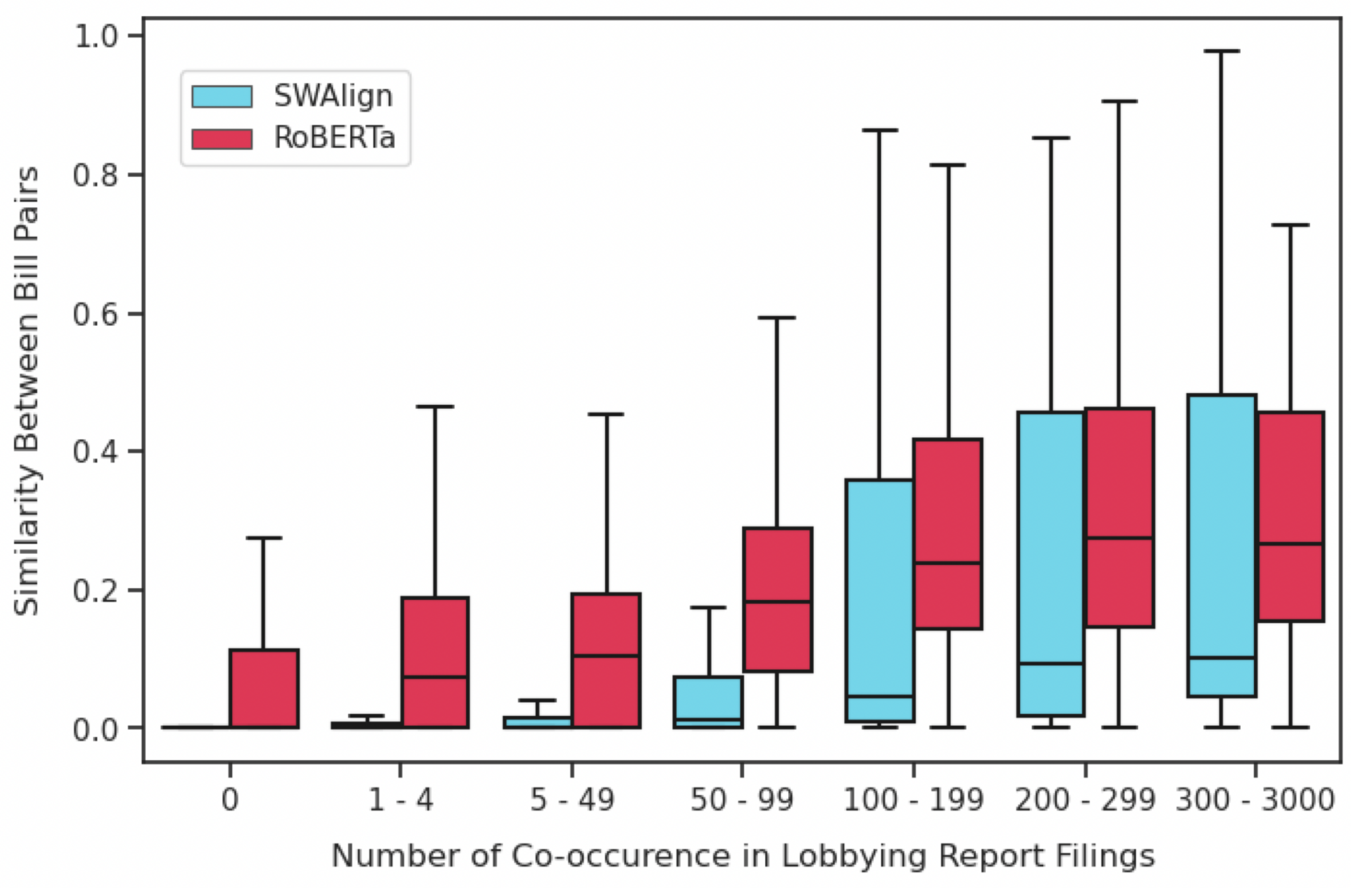}
\caption[]{Bill similarity result.}
\label{application:max_bill_similarity}
\end{figure}

Next, we group pairs of bills into 7 distinct categories according to the level of co-occurrence to conduct a conceptually cleaner analysis. Table \ref{application:bill_pair_cooccurrence} in Appendix \ref{appendix:bill_analysis} presents the distribution of the co-occurence across these categories.  Note that there is a big difference in the average number of subsections in the bill for each category: The average number of subsections of the bills of \textit{300-3000} type is over 500, whereas for \textit{1-4} type, it is less than 60. This is consistent with the fact that longer bills tend to reflect diverse interests and hence lobbied by many interest groups. To make a fair comparison in similarity, we filter out the bill pairs with less than 10 subsections. For range 0, i.e., a pair of bills that have never been reported together, we randomly sample bill pairs from ~\congress~session. For each type, we select up to 600 samples, which yields a total of 2,718 bill pairs for our analysis.

\paragraph{Measuring Similarity Method}
We aggregate subsection-level classification results into a bill-level similarity as follows.
We denote bill $B_i$ with $n$ subsections $s$ by $B_i = \{s_{i1}, \ldots, s_{in}\} \in \mathbb{B}$ , where $\mathbb{B}$ is a set of bills. $\mathcal{B}$ is a set of bill pairs, $\mathcal{B} = \{(B_i, B_j)|B_i \in \mathbb{B}, B_j \in \mathbb{B}, i \neq j\}$. 
For any given two bills $B_i$ and $B_j$, we develop two measures of similarity, $\sigma_{ij}$ and $\sigma_{ji}$, by switching the order of computation. Specifically, we begin by considering the similarity between each subsection $s_{ik}$ of $B_i$ and each subsection $s_{jk^\prime}$ of $B_j$ according to the mapping $f$ that we developed earlier in Section~\ref{sec:measure}, i.e., $f: \{s_{ik}, s_{jk^\prime}\} \rightarrow \{0,1,2,3,4\}$. We then take the sum of the maximum similarity measures across each combination of subsection pairs in order to get the similarity between two bills:$\sigma_{ij} \equiv \sum_{s_{ik} \in B_i} \max\{f(s_{ik} , s_{jk^\prime})\}, \forall s_{jk^\prime} \in B_j$. $\sigma_{ji}$ is defined similarly. We then normalize the similarity measures so that the measure lies between 0 and 1: $\sigma^\ast_{ij} = \frac{\sigma_{ij}}{4 \cdot |B_i|}$ and $\sigma^\ast_{ji} = \frac{\sigma_{ji}}{4 \cdot |B_j|}$. Finally, we take the maximum of the two measures to get our final bill-level similarity: $\max\{\sigma^\ast_{ij}, \sigma^\ast_{ji}\}$. Note that we do not take the average, because, for a bill pair with large differences in the number of subsections, the average similarity will be arbitrarily small even when two bills share highly similar contents across a few subsections.

\paragraph{Bill-level Results}
Figure~\ref{application:max_bill_similarity} and Table \ref{application:max_bill_average_median} in Appendix \ref{appendix:bill_analysis} summarize our findings. As expected, the similarity tends to increase as the co-occurrence frequency increases. This suggests the proposed similarity measure accurately captures the similarity between two bills. Consistent with our earlier findings, we find that RoBERTa shows higher performance than SWAlign in terms of the mean and median of the predicted bill-level similarities. Finally, the distributions of the prediction between SWAlign and RoBERTa are also significantly different (p-value from t-test: 7.941e-81 < 0.001, ***). Our novel measurement at the bill-level opens a new avenue for empirical research of complex political dynamics as politicians vote on, introduce, and amend legislative bills.

\section{Conclusion}
In this paper, we developed a methodology to identify the links between legislative bills. We proposed 5-class classification task and constructed a human-labeled dataset of 4,721 legislative text pairs. Additionally, we generated synthetic datasets varying the degree of similarities to boost the classification performance.
Furthermore, we applied the models trained on the subsection-level to section- and bill-level pairs. We have shown that our approach successfully captures similarities across legislative bills at different levels of aggregation.

Our model provides an empirical framework as well as novel measurements that allow scholars to investigate bill-to-bill networks in legislative politics. This is important as policy preferences of various political actors are reflected in different parts of bills at different stages in the legislative process. Bill networks can be a medium that helps scholars both to understand the legislative process and to analyze the hidden connections present between various actors such as politicians, lobbyists, and interest groups.
For instance, using our model, researchers can conduct an in-depth analysis of how several seemingly different bills, in effect, originate from an identical policy idea and political interest \citep{wilkerson2015tracing}. 
Moreover, our model has broader applications in studying legal texts. In particular, it can be used to study policy diffusion across states \citep{linder2020text, burgess2016legislative} or across countries \citep{pagliari2020exploring, james2021internationalization}, find lobbying patterns between politicians and lobbyists \citep{mckay2018fundraising}, and trace constantly changing relevant laws or regulations  \citep{chalkidis-etal-2021-regulatory}.

Much research is needed to understand the political process that underlies the bill linkages. While we have focused on the measurements, we hope that the outputs that we produced in this paper could be useful not only for improving methodological research but also for future computational social science research examining political connections in the legislative process that are often hidden to public eyes.

\section{Ethical Consideration}
Among 9 annotators, the 7 annotators excluding two authors are compensated properly for their work according to the standards set by the two academic institutions that the authors are affiliated with.

\section*{Acknowledgements}
This work was supported by the National Research Foundation of Korea (NRF) grant funded by the Korea government (MSIT) (2019R1A2C1085371) and MIT International Science and Technology Initiatives (MISTI) Global Seed Fund. In Song Kim acknowledges financial support from the National Science Foundation (SES-2017315). The authors would also like to thank the annotators for their help, the anonymous reviewers, and Donghoon Han for their thoughtful comments and suggestions.


\bibliography{anthology,custom}
\bibliographystyle{acl_natbib}

\clearpage
\appendix
\section{Appendix}
\label{sec:appendix}
In Appendix, we show the additional results of experiment, the detailed model settings, boilerplate description, and coding rules used for building the human-labeled subsection pair dataset. 

\section{Analysis of Subsection Pair Similarity and Model's Prediction}
\label{appendix:extra_annotation_experiment}
The similarity distribution of subsection pairs is highly skewed, i.e., it is rare to observe that any random subsection pairs are related. We choose the 0.85 threshold to reflect the skewed distribution after examining sample pairs across all similarity ranges. To verify whether our model performs outside of the similarity range, we conduct an extra experiment by randomly sampling 2,468 pairs with similarity scores less than 0.85. As expected, our model predicts most of the pairs as \textit{Unrelated}.
\begin{itemize}
\item \textit{Unrelated}: 98.87\% (2,440 pairs) 
\item \textit{Partially Related}: 1.01\% (25 pairs) 
\item \textit{Related}: 0.12\% (3 pairs)
\end{itemize}
Furthermore, we annotate an additional 213 pairs that are randomly sampled from these 2,468 pairs, including the pairs predicted as \textit{Related} or \textit{Partially Related}. All pairs are judged by an annotator to be \textit{Unrelated}, except for two boilerplate pairs. This result shows that our model works on pairs in all similarity ranges.

\section{Additional Results}
\label{appendix:result}
\paragraph{Ablation Study of Synthetic Dataset} Table ~\ref{appendix:result_ablation_val} is validation result of ablation study for augmentation operations. The result shows that applying both swap and synonym operations helps increase the classification performance across all BERT models.

\begin{table}[!htbp]
\centering
\tiny
\begin{tabular}{@{}l|cc|cc|cc@{}}
\toprule
          & \multicolumn{2}{c|}{RoBERTa}   & \multicolumn{2}{c|}{Legal-BERT} & \multicolumn{2}{c}{BERT}      \\
operation & Acc.          & F1            & Acc.           & F1            & Acc.          & F1            \\  \midrule
swap      & 69.8$_{3.5}$          & 62.3$_{3.9}$          & 61.3$_{2.0}$           & 55.2$_{1.8}$          & 57.9$_{3.8}$          & 51.9$_{3.3}$          \\
synonym   & 69.3$_{2.3}$          & 63.7$_{2.1}$          & 63.3$_{1.8}$           & \textbf{58.4}$_{2.2}$ & 58.1$_{3.0}$          & 54.0$_{2.9}$          \\
both      & \textbf{72.8}$_{2.7}$ & \textbf{64.5}$_{2.4}$ & \textbf{64.4}$_{3.0}$  & \textbf{58.4}$_{2.9}$ & \textbf{58.5}$_{2.3}$ & \textbf{54.7}$_{2.2}$ \\ \bottomrule
\end{tabular}
\caption{ Ablation  study  of validation set. The augmentation  operations are synonym  and  swap.  Results  are  from  models  trained on 10,000 synthetic pairs.}
\label{appendix:result_ablation_val}
\end{table}

\paragraph{Effectiveness of Synthetic Dataset} Table ~\ref{appendix:result_f1_score_class_val} shows the performance of each relation class for validation dataset.
As the table suggests, the result is consistent with the testset. Training on both synthetic and human-labeled pairs boost the classification score across all the relation classes.
\begin{table}[!htbp]
\centering
\small
\begin{tabular}{@{}l|ccc@{}}
\toprule
Class               & Synthetic & Human & Syn.+Human \\ \midrule
4 Identical         & 90.6$_{3.5}$      & 96.0$_{1.2}$        & \textbf{96.4}$_{0.4}$   \\
3 Almost Identical  & 58.1$_{6.8}$      & 84.5$_{10.5}$        & \textbf{86.1}$_{5.2}$   \\
2 Related           & 63.2$_{6.0}$      & 76.1$_{3.1}$        & \textbf{79.4}$_{1.7}$   \\
1 Partially Related & 27.6$_{5.7}$      & 50.4$_{2.0}$        & \textbf{52.3}$_{0.8}$   \\
0 Unrelated         & 83.0$_{2.4}$      & 94.9$_{0.9}$        & \textbf{95.7}$_{0.4}$   \\ \midrule
Accuracy            & 72.8$_{2.7}$      & 88.3$_{0.9}$        & \textbf{89.4}$_{0.4}$   \\
macro F1            & 64.5$_{2.4}$      & 80.4$_{2.5}$        & \textbf{82.0}$_{1.0}$   \\ \bottomrule
\end{tabular}
\caption[]{F1 of each class, average accuracy and average F1 for validation set. We experiment with three different training data types: synthetic, human-labeled, synthetic+human-labeled dataset. We report results generated from RoBERTa.}
\label{appendix:result_f1_score_class_val}
\end{table}

\begin{table*}[!htbp]
\centering
\small
\begin{tabular}{@{}rc|cc|cc|cc|cc@{}}
\toprule
\multicolumn{2}{c|}{Dataset} & \multicolumn{2}{c|}{RoBERTa} & \multicolumn{2}{c|}{Legal-BERT} & \multicolumn{2}{c|}{BERT} & \multicolumn{2}{c}{SWAlign} \\
Syn.     & Hu.      & Acc.         & F1           & Acc.           & F1            & Acc.        & F1         & Acc.         & F1           \\  \midrule
0K              & 3305       & 88.3         & 80.4         & 89.4           & 81.7          & 87.1        & 77.6       & 66.9         & 32.1         \\  \midrule
1K           & 3305       & 88.6         & 81.0         & 88.1           & 79.8          & 86.7        & 77.0       & 63.1         & 40.0         \\
5K           & 3305       & 89.2         & 81.5         & 89.2           & 82.0          & 88.0        & 79.3       & 62.1         & 39.5         \\
10K          & 3305       & 89.4         & 82.0         & 89.5           & 82.0          & 87.8        & 79.2       & 61.0         & 39.5         \\
50K          & 3305       & 89.1         & 81.8         & 89.0           & 81.3          & 87.8        & 79.6       & 56.6         & 37.9         \\
100K         & 3305       & 88.4         & 80.9         & 89.3           & 81.4          & 88.5        & 80.7       & 53.4         & 36.6         \\ \bottomrule
\end{tabular}
\caption{Validation results with various sizes of synthetic and human-labeled data for all models.}
\label{appendix:classification_val}
\end{table*}

\begin{table*}[!htbp]
\centering
\small
\begin{tabular}{@{}rc|cc|cc|cc|cc@{}}
\toprule
\multicolumn{2}{c|}{Dataset} & \multicolumn{2}{c|}{RoBERTa} & \multicolumn{2}{c|}{Legal-BERT} & \multicolumn{2}{c|}{BERT} & \multicolumn{2}{c}{SWAlign} \\
Syn.     & Hu.      & Acc.         & F1           & Acc.           & F1            & Acc.        & F1         & Acc.         & F1           \\  \midrule
0K              & 3305       & 86.9         & 76.8         & 87.5           & 76.9          & 85.3        & 73.7       & 67.5         & 32.4         \\  \midrule
1K           & 3305       & 87.4         & 77.9         & 87.5           & 77.1          & 85.1        & 74.0       & 62.2         & 37.0         \\
5K           & 3305       & 88.6         & 79.6         & 88.1           & 78.7          & 86.5        & 76.1       & 61.0         & 37.0         \\
10K          & 3305       & 88.9         & 79.9         & 88.2           & 78.8          & 85.8        & 75.1       & 60.3         & 37.3         \\
50K          & 3305       & 88.6         & 79.4         & 88.2           & 79.0          & 86.4        & 76.3       & 56.7         & 36.0         \\
100K         & 3305       & 88.6         & 79.4         & 87.8           & 78.4          & 86.7        & 76.4       & 53.9         & 36.2         \\ \bottomrule
\end{tabular}
\caption{Test results with various sizes of synthetic and human-labeled data for all models.}
\label{appendix:classification_test}
\end{table*}

\paragraph{Performance for Different Size of Human-labeled and Synthetic Dataset} Table~\ref{appendix:classification_val} and Table~\ref{appendix:classification_test} show the classification performance with various sizes of synthetic and human-labeled data for validation and testdataset, respectively. We apply two-stage training for all models. Training models on synthetic datasets tend to show higher performance than training on human-labeled pairs only. RoBERTa shows the best performance among all models. It achieves the best F1 79.9 and accuracy 88.9 trained on 10,000 synthetic pairs and all 3,305 human-annotated pairs, followed by Legal-BERT (F1: 79.0, accuracy: 88.2), BERT (F1: 86.7, accuracy: 76.4) and SWAlign (F1: 37.3, accuracy: 60.3).  The result demonstrates the effectiveness of synthetic dataset across variant models.

\begin{table*}[!htbp]
\centering
\small
\begin{tabular}{@{}rr|cc|cc|cc|cc@{}}
\toprule
\multicolumn{2}{c|}{Dataset}                               & \multicolumn{2}{c|}{RoBERTa} & \multicolumn{2}{c|}{Legal-BERT} & \multicolumn{2}{c|}{BERT} & \multicolumn{2}{c}{SWAlign} \\
\multicolumn{1}{c}{Synthetic} & \multicolumn{1}{c|}{Human} & lr             & Epoch      & lr              & Epoch        & lr           & Epoch     & Norm       & C              \\  \midrule
10K                           & 0                         & 2e-5         & 3            & 2e-5          & 2              & 3e-5       & 4           & l2           & 1e-2         \\
0                             & 3305                      & 2e-5         & 4            & 1e-5          & 4              & 2e-5       & 4           & l2           & 1            \\
10K                           & 3305                      & 2e-5         & 4            & 1e-5          & 4              & 2e-5       & 4           & l2           & 1e+2         \\ \bottomrule
\end{tabular}
\caption{The best parameters of models. lr indicates a learning rate.}
\label{appendix:model_params}
\end{table*}

\section{Model Settings}
\label{appendix:model_settings}
We use four GeForce GTX 1080 Ti for training a model, where the runtime varies depending on model and the training set size. Training all 3305 human-labeled pairs takes about 7 minutes. 
Training 10,000 synthetic pairs with 4 epochs takes about 40 minutes.
For SWAlign, we tune the hyper-parameters of the logistic regression with the following range: C $\in \{1e-3,1e-2,1e-1, 1, 1e+1,1e+2,1e+3\}$, which is inverse of regularization strength, and Norm $ \in \{l2, none\}$.
The best F1 parameters are depending on training settings, and we report the best score of optimal parameters in Table~\ref{appendix:model_params}.
We evaluate our model's accuracy and F1 score with the scikit-learn library \footnote{\url{https://scikit-learn.org/stable/modules/generated/sklearn.metrics.classification_report.html}}.

\begin{table}[!htbp]
\small
\centering
\begin{tabular}{@{}l|r|r@{}}
\toprule
\# Co-occurrence & Total  & \# Selected Pairs   \\ \midrule
300 - 3000       & 48     & 41   \\
200 - 299        & 71     & 49   \\
100 - 199        & 371    & 243  \\
50 - 99          & 1,140   & 585  \\
5 - 49           & 80,511  & 600  \\
1 - 4            & 136,996 & 600  \\
0                & -      & 600  \\ \midrule
Sum              & 2,19,137 & 2,718 \\ \bottomrule
\end{tabular}
\caption[]{Statistics of bill pairs appearing together in lobby reports. We use this bill-level related pairs as for measuring bill-level similarity.}
\label{application:bill_pair_cooccurrence}
\end{table}

\begin{table}[!htbp]
\small
\centering
\begin{tabular}{@{}l|rr|rr@{}}
\toprule
         & \multicolumn{2}{c|}{SWAlign} & \multicolumn{2}{c}{RoBERTa} \\
Range    & Average   & Median  & Average   & Median   \\ \midrule
300-3000 & 24.3      & 10.1  & 33.2      & 26.8  \\
200-299  & 25.4      & 9.4   & 33.1      & 27.4 \\
100-199  & 20.0      & 4.6   & 31.5      & 24.0 \\
50-99    & 10.9      & 1.3   & 22.6      & 18.2 \\
5-49     & 3.6       & 0.0   & 13.3      & 10.5 \\
1-4      & 1.4       & 0.0   & 11.0      & 7.5 \\
0        & 0.2       & 0.0   & 6.0       & 0.0 \\ \bottomrule
\end{tabular}
\caption[]{Average and median of bill similarity for each range.}
\label{application:max_bill_average_median}
\end{table}

\section{Bill-level Analysis}
\label{appendix:bill_analysis}
\paragraph{Bill-level Pairs} Table~\ref{application:bill_pair_cooccurrence} shows the number of total and sampled bill pairs for each range used for measuring bill similarity.

\paragraph{Bill-level Results} Table~\ref{application:max_bill_average_median} shows the average and median of bill similarity score for ranges. The similarity tends to increase as the co-occurrence frequency increases. This suggests the proposed similarity measure accurately captures the similarity between two bills. For both average and median, RoBERTa gets the higher scores than SWAlign on predicting bill-level similarities.

\section{Boilerplate}
\label{appendix:boilerplate}
Two subsections share common templated text that is repeatedly used in bills. These templates create an arbitrary similarity between bills that are otherwise unrelated in policy.  Since a boilerplate pair is likely to be labeled as a  \textit{Identical} or \textit{Almost Identical} class, annotators are instructed to label boilerplate so that these data instances can be filtered out from the dataset during preprocessing. 12 subsection titles are involved in Boilerplate. Table~\ref{appendix:boilerplate_example} shows the boilerplate examples for each  subsection's title.

\begin{table*}[]
\centering
\small
\begin{tabular}{@{}l|l@{}}
\toprule
Subsection Title                  & Example                                                                                                                                                                                                                                                                                                                                                                                                                                                                                                                                                                                                                                                                                                      \\ \midrule
Effective Date                  & \begin{tabular}[c]{@{}l@{}}The amendments made by subsection (a) apply to goods entered, or \\
withdrawn from warehouse for consumption, on or after  the 15th day \\
after the date of the enactment of this Act.\end{tabular} \\ \midrule
Authorization of Appropriations & \begin{tabular}[c]{@{}l@{}}There are authorized to be appropriated such sums as are necessary \\
to carry out this title. Any amount appropriated under this subsection \\
shall remain available, without fiscal year limitation, until expended.\end{tabular} \\ \midrule
Vacancies                       & \begin{tabular}[c]{@{}l@{}}(1) In general.--The Board shall fill any vacancy in the membership \\
of the Council in the same manner as the original appointment. \\
The vacancy shall not affect the power of the remaining members \\
to execute the duties of the Council. \\
(2) Vacancy appointments.--Any member appointed to fill a vacancy \\
shall serve for the remainder of the term for which the predecessor \\
of the member was appointed.\\
(3) Reappointment.--The Board may reappoint an appointed member of the \\
Council for a second term in the same manner as the original appointment. \end{tabular} \\  \midrule
Termination                     & \begin{tabular}[c]{@{}l@{}}The program shall terminate, and no loan may be made under this section,\\ 
on or after the date that is 25 years after the date on which amounts are\\
initially appropriated under subsection (e).\end{tabular}                                                                                                                                                                                                                                                                                                                                                                                                                                                               \\ \midrule
Table of Contents               & \begin{tabular}[c]{@{}l@{}}The table of contents for this Act is as \\ follows:\\ \\ Sec. 1. Short title; table of contents.\\ Sec. 2. Findings; policy.\\ Sec. 3. Definitions.\end{tabular}                                                                                                                                                                                                                                                                                                                                                                                                                                                                                                                   \\ \midrule
Short Title                     & \begin{tabular}[c]{@{}l@{}}This Act may be cited as the ``Small Business Innovation to Job \\ Creation Act of 2010''.\end{tabular}                                                                                                                                                                                                                                                                                                                                                                                                                                                                                                                                                                             \\ \midrule
Reference                       & \begin{tabular}[c]{@{}l@{}}Except as otherwise expressly provided, whenever in this title an amendment\\
or repeal is expressed in terms of an amendment to, or repeal of, a section \\
or other provision, the reference shall be considered to be made to a section \\
or other provision of the Internal Revenue Code of 1986. \end{tabular}                                                                                                                                                                                                                                                                                                                                                              \\ \midrule
Sunset                          & \begin{tabular}[c]{@{}l@{}}This title, and any amendments made by this title are repealed, and the \\
requirements under this title shall terminate, on December 31, 2012.\end{tabular}                                                                                                                                                                                                                                                                                                                                                                                                                                                                                                                     \\ \midrule
Appropriation                   & \begin{tabular}[c]{@{}l@{}}There are authorized to be appropriated to the Secretary of State \\ \$ 6 , 000 , 000 for fiscal year 2010 to carry out this section .\end{tabular}                                                                                                                                                                                                                                                                                                                                                                                                                                                                                                                                 \\  \midrule
Severability                    & \begin{tabular}[c]{@{}l@{}}If any provision of this Act or any amendment made by this Act, or \\ the application of a provision or amendment to any person or \\ circumstance, is held to be unconstitutional, the remainder of this Act \\ and the amendments made by this Act, and the application of the \\ provisions and amendments to any person or circumstance, shall not be \\ affected by the holding.\end{tabular}                                                                                                                                                                                                                \\  \midrule
Matching Requirement            & \begin{tabular}[c]{@{}l@{}}An eligible municipality seeking a grant under this section for a community\\
greening initiative shall agree to make available non-Federal funds to carry \\
out the community greening initiative in an amount equal to not less than 50 \\
percent of the grant awarded to such eligible municipality under this section.\end{tabular}                                                                                                                                                                                                                                                                                                                                  \\  \midrule
Definitions                     & \begin{tabular}[c]{@{}l@{}}In this section--\\
(1) the term ``Administrator'' means the Administrator of the Small Business \\Administration; \\
(2) the terms ``Federal agency'', ``research'', and ``Small Business Innovation\\
Research Program'' have the meanings given such terms in section 9(e) of the \\
Small Business Act (15 U.S.C.638(e)); and\\
(3) the term ``second phase'' means the phase described in section 9(e)(4)(B) \\
of the Small Business Act (15 U.S.C.638(e)(4)(B)).\end{tabular}                                                                                                    \\ \bottomrule
\end{tabular}
\caption[]{Boilerplate Examples of 12 subsection titles.}
\label{appendix:boilerplate_example}
\end{table*}



\section{Coding Rules}
\label{appendix:coding_rules}
We provide description of 5-relation types to coders when building human-labeled pairs.

\paragraph{Subsection Relation Types}Table~\ref{appendix:desciprtion_coding_rules} shows the description of coding rules for relation classes. Additionally, we ask coders to mark subsection pairs, which are included in Boilerplate or templates texts. The pairs marked with one of those, we filter out them in preprocessing step.

\paragraph{Example of Common Phrases in Bills} Table~\ref{appendix:n_gram_samples} is samples of the common phrases appear in bill text. As for reference, we provide top 50 common N-gram list to coders for annotation.  We set N from 3 to 10.
\begin{table*}[!htbp]
\small
\centering
\begin{tabular}{@{}l|l@{}}
\toprule
Relation Type             & Description                                                                                                                                                                                                                                                                                         \\ \midrule
{[}4{]} Identical         & copy and paste                                                                                                                                                                                                                                                                                   \\ \midrule
{[}3{]} Almost Identical  & \begin{tabular}[c]{@{}l@{}}•  word (object) level change\\ 
\hphantom{~~} – only single number is different (bill number)\\ •  paraphrasing: same meaning without adding / deleting object or political idea\\ 
\hphantom{~~} – if a new idea is added, the pair should be classified as class ``[2] Related''\\
\hphantom{~~} – only order change\end{tabular} \\ \midrule
{[}2{]} Related           & \begin{tabular}[c]{@{}l@{}}•  at least about 50\% of political phrase are overlapped\\ • new specific political idea is added / deleted\\ • if words are added / deleted involving extra meaning (different idea), this pair likely \\    to be considered class ``[2] Related'' \end{tabular}                 \\ \midrule
{[}1{]} Partially Related & •  less than 50\% of political phrases / clauses are overlapped                                                                                                                                                                                                                                     \\ \midrule
{[}0{]} Unrelated         & \begin{tabular}[c]{@{}l@{}}• different / unrelated section\\
\hphantom{~~} – two subsections have no common sentence (phrase / clause), or all common texts \\  
\hphantom{~~}   are included in top common N-grams list (Table~\ref{appendix:n_gram_samples} in Appendix \ref{appendix:coding_rules})\end{tabular} \\ \midrule \midrule

{[}B{]} Boilerplate    & \begin{tabular}[c]{@{}l@{}}these will be common templated sections (i.e coin bills, visa sections, etc) that relate \\ to other bills due to the similarity of common structure and not because of their actual \\ similarity in intent.\end{tabular} \\ \midrule
{[}E{]} Excluded       & \begin{tabular}[c]{@{}l@{}}text which does not specify specific political ideas or unrelated object to subsection \\ contents (e.g., Definition, Tables)\end{tabular}                                                                               \\ \midrule
Text to ignore & \begin{tabular}[c]{@{}l@{}}•  top 50 common Ngram (Table~\ref{appendix:n_gram_samples} in Appendix \ref{appendix:coding_rules})\\ •  apply to all classes\end{tabular} \\
\bottomrule
\end{tabular}
\caption[]{Description of Coding Rules. This table provides a further description of our coding rule. We note that one of key distinctions between ``[3] Almost Identical'' and ``[2] Related'' subsection relation type is that the former concerns two subsections with the same policy idea, while the latter characterizes the subsection pairs with different policy ideas. This is capture by the following coding rule for ``[2] Related'' type: ``a new specific political idea is added / deleted''.}
\label{appendix:desciprtion_coding_rules}
\end{table*}

\begin{table*}[!htbp]
\small
\centering
\begin{tabular}{@{}r|l|r@{}}
\toprule
N  & N-gram                                                & Count \\ \midrule
 & is amended by adding at the end the following new     & 5174  \\
 & after the date of the enactment of this Act the       & 2540  \\
 & days after the date of the enactment of this Act      & 1976  \\
10 & Code is amended by adding at the end the following    & 1764  \\
 & of the Internal Revenue Code of 1986 is amended by    & 1662  \\
 & the date of the enactment of this Act the Secretary   & 1540  \\
 & after the date of enactment of this Act the Secretary & 1529  \\ \midrule
  & the Internal Revenue Code of 1986 is amended by       & 1666  \\
  & the date of enactment of this Act the Secretary       & 1580  \\
  & date of the enactment of this Act the Secretary       & 1544  \\
9  & the Public Health Service Act 42 U S C                & 1542  \\
  & of the Public Health Service Act 42 U S               & 1520  \\
  & Not later than 180 days after the date of             & 1478  \\ \midrule
  & 2 by adding at the end the following                  & 1420  \\
  & year after the date of enactment of this              & 1407  \\
8  & 1 year after the date of enactment of                 & 1380  \\
  & and 2 by adding at the end the                        & 1333  \\
  & United States Code is amended by adding at            & 1273  \\
  & States Code is amended by adding at the               & 1273  \\ \midrule
  & year after the date of enactment of                   & 1728  \\
  & Revenue Code of 1986 is amended by                    & 1676  \\
7  & Not later than 180 days after the                     & 1672  \\
  & of enactment of this Act the Secretary                & 1588  \\
  & enactment of this Act the Secretary of                & 1585  \\ \midrule
  & days after the date of enactment                      & 2453  \\
  & of such Code is amended by                            & 2419  \\
  & the end the following new paragraph                   & 2332  \\
6  & later than 180 days after the                         & 2170  \\
  & striking the period at the end                        & 2161  \\
  & 1 year after the date of                              & 2141  \\
  & the end the following new subsection                  & 1922  \\ \midrule
  & at the end the following                              & 17303 \\
5  & by adding at the end                                  & 17116 \\
  & adding at the end the                                 & 16603 \\
  & the date of enactment of                              & 11070 \\ \midrule
  & the end the following                                 & 17326 \\
  & at the end the                                        & 17310 \\
4  & adding at the end                                     & 17171 \\
  & by adding at the                                      & 17122 \\
  & after the date of                                     & 16799 \\
  & date of enactment of                                  & 11129 \\ \midrule
  & at the end                                            & 25197 \\
  & with respect to                                       & 24365 \\
3  & under this section                                    & 23672 \\
  & of this Act                                           & 20146 \\
  & after the date                                        & 19807 \\
  & 1 In general                                          & 17316 \\ \bottomrule
\end{tabular}
\caption[]{Samples of top 50 N-gram in bill text.}
\label{appendix:n_gram_samples}
\end{table*}

\end{document}